# Prediction of Electric Multiple Unit Fleet Size Based on Convolutional Neural Network

Boliang Lin *

School of Traffic and Transportation, Beijing Jiaotong University, Beijing 100044, China

**Abstract:** With the expansion of high-speed railway network and growth of passenger transportation demands, the fleet size of electric multiple unit (EMU) in China needs to be adjusted accordingly. Generally, an EMU train costs tens of millions of dollars which constitutes a significant portion of capital investment. Thus, the prediction of EMU fleet size has attracted increasing attention from associated railway departments. First, this paper introduces a typical architecture of convolutional neural network (CNN) and its basic theory. Then, some data of nine indices, such as passenger traffic volume and length of high-speed railways in operation, is collected and preprocessed. Next, a CNN and a backpropagation neural network (BPNN) are constructed and trained aiming to predict EMU fleet size in the following years. The differences and performances of these two networks in computation experiments are analyzed in-depth. The results indicate that the CNN is superior to the BPNN both in generalization ability and fitting accuracy, and CNN can serve as an aid in EMU fleet size prediction.

## 1. Introduction

Nowadays, EMU trains have greatly improved the passenger transportation efficiency due to their high speed and convenient transfer with other transportation modes. There are different types of EMU trains in operation at present, e.g., China Railway High-speed (CRH, China), train à grande vitesse (TGV, France), Inter City Express (ICE, Germany) and Shinkansen N700 Series (Japan). According to the Medium and Long Term Railway Network Plan issued in 2016, the total length of high-speed railway in operation in China will reach 30000 km in 2020, which is merely 22000 km in 2016. As an important means of passenger transportation, the fleet size of EMU has to match the expansion of high-speed railway network, and satisfy the increasing passenger transportation demands. It should be noted that lack of EMU might result in a waste of railway network capacity and preclude the delivery of passenger flows. In contrast, excessive amount of EMU might lead to the problem of deadheading equipments and idling crews. Additionally, the purchasing cost of an EMU train consisting of eight EMUs generally exceeds tens of millions of dollars, which constitutes a significant portion of capital investment. Thus, the prediction of EMU fleet size is of great value in practice and has attracted increasing attention from associated railway departments. This paper aims to predict EMU fleet size scientifically and provide a solid support for associated departments in decision making.

There is a body of literature devoted to EMU maintenance and lifetime, as well as railway-induced vibrations, whereas, few of them focus on EMU fleet size prediction. Li et al. (2013) proposed an approach for predicting annual workload of major maintenance of EMU. Cai et al. (2015) constructed a new forecasting method for operation safety of high-speed train. Chen et al. (2017) presented a prediction algorithm for harmonic current based on confidence intervals. The prediction of railway-induced vibrations was studied by Yang et al. (2015), Sun et al. (2016), Hou et al. (2016) and Kouroussis et al. (2016). Pugi et al. (2015) presented the development of a modular tool for the prediction of train braking performance and devoted to the accurate prediction of stopping distances. Lestoille et al.

---

* Corresponding author. Email: bllin@bjtu.edu.cn

(2016) proposed a stochastic predictive model for predicting the statistical quantities of a vector-valued random indicator related to the nonlinear dynamic responses of the high-speed train excited by stochastic track irregularities.

Researchers generally adopt Grey Model, BPNN and combination forecasting methods in prediction, whereas quite a few implement CNN. Since the concept of regional receptive field was proposed by Hubel and Wiesel in 1962, plenty of researchers have conducted extensive and in-depth research of CNN in the fields of image identification and speech recognition. Fukushima and Miyake (1982) presented the concept of neocognitron on the basis of receptive field. Then, a CNN aiming at document recognition was developed by Lécun et al. (1998), the design of which combined regional receptive field, weight sharing and down-sampling, ensuring the translation invariance to some extent. Later, Ji et al. (2013) established a novel 3D CNN model for action recognition, extracting features from both the spatial and the temporal dimensions by performing 3D convolutions. Brachmann and Redies (2016) proposed a method for measuring symmetry in images by using filter responses from CNN. Zhang and Mu (2017) put forward an efficient technique involving Multiple Scale Faster Region-based Convolutional Neural Networks (Faster R-CNN) to detect ears from 2D profile images. Zhang et al. (2017) explored an automatic radar waveform recognition system to detect, track and locate the low probability of intercept radars and proposed a hybrid classifier combining convolutional neural network and Elman neural network. Sainath et al. (2013) explored to applying CNNs to large vocabulary speech tasks. Similarly, Abdel-Hamid et al. (2014) proposed a limited-weight-sharing scheme for speech recognition. Karpathy et al. (2014) provided an extensive empirical evaluation of CNNs on large-scale video classification. Recently, researchers have further improved its performance and proposed new structure of CNN, e.g., Visual Geometry Group (VGG) of Oxford University (Simonyan et al., 2014), GoogLeNet of Google (Szegedy et al., 2015), and ResNet from Microsoft (He et al., 2015). Currently, a research trend of CNN is combining traditional algorithms.

The paper is organized as follows: Section 2 describes the typical architecture and basic theory of CNN. In Section 3, we analyze and normalize the data of nine indices. Then, computational experiments of predicting EMU fleet size based on CNN and BPNN are carried out in Section 4 and 5, respectively. Conclusions are presented in Section 6.

## 2. Methodology

### 2.1 Typical Architecture of CNN

CNN is a type of feed-forward deep neural network with convolutional structure (Chang et al., 2016), i.e., information can only be transmitted from the input end to the output end. Typically, CNN consists of a stack of distinct layers, including input layer, convolutional layer, pooling layer, fully connected layer as well as loss layer. Compared with common neural networks, CNN features convolutional layer and pooling layer. The convolutional layer is involved with a set of filters (also called convolutional kernels). Each neuron of this layer is connected to only a small region of the input volume, and the extent of this connectivity is called the receptive field of the neuron. The pooling layer serves to progressively reduce the number of parameter and amount of computation. There are several non-linear functions to implement pooling among which max pooling is the most common. Moreover, the pooling operation provides a form of translation invariance. The typical architecture of CNN is depicted in Fig. 1.

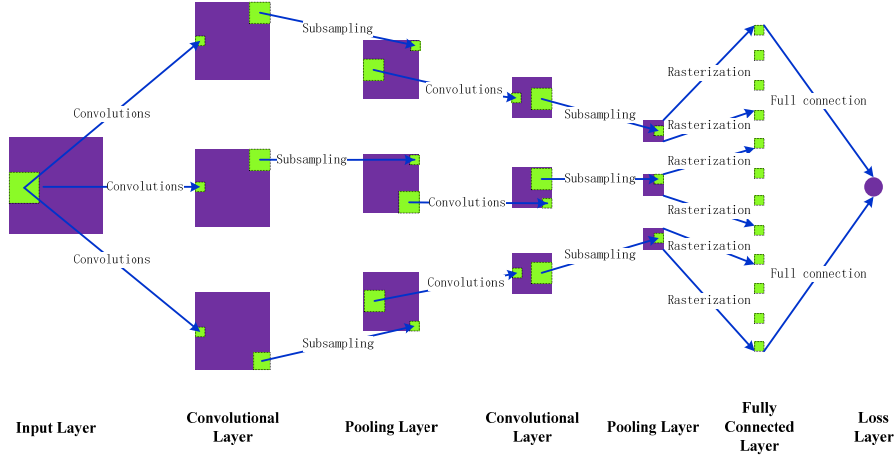

Fig. 1. Typical architecture of CNN

## 2.2 Basic Theory of CNN

The process of prediction applying CNN can be divided into three stages, including network construction, training and prediction (Li et al., 2016). The construction stage involves the design of network architecture, determination of convolutional kernel and pooling filter, choice of non-linear function and loss function, initialization of parameters, etc. The training stage refers to the forward propagation of training samples, backpropagation of error and adjustment of parameters, aiming at minimizing the deviation between the predicted value and target value. The prediction stage is involved with the forward propagation of predicting samples over the trained network and the generation of predicted value.

Generally speaking, the output feature map of layer $L_k$ is exactly the input feature map of layer $L_{k+1}$. If $L_k$ is a convolutional layer, then the output feature map $y^{j(k)}$ can be obtained based on the following equation:

$$y^{j(k)} = f_k\left(\sum_{i(k)} x^{i(k)} * w_{j(k)}^{i(k)} + b^{j(k)}\right) \qquad (1)$$

where $i(k)$ and $j(k)$ are ordinal numbers of input feature map $x^{i(k)}$ and output feature map $y^{j(k)}$ in layer $L_k$, respectively. $w_{j(k)}^{i(k)}$ represents the convolutional kernel corresponding to $x^{i(k)}$ and $y^{j(k)}$. $*$ is the notation of convolution, which refers to that a kernel is convolved across the width and height of the input feature map with a given stride. During the forward pass, we calculate the dot product of the kernel and the input feature map. If $x^{i(k)}$ has a size of $r \times r$, and $w_{j(k)}^{i(k)}$ has a size of $c \times c$, then the size of output feature map $y^{j(k)}$ will be equal to $(r-c+1) \times (r-c+1)$. Once the convolution process is completed, bias $b^{j(k)}$ will be added, and non-linear function $f_k(x)$ will be subsequently applied. Typically, Rectified Linear Units (ReLU), hyperbolic tangent (tanh) and sigmoid function (sigmoid) are three major non-linear functions.

Similarly, if $L_k$ is a pooling layer, then the output feature map $y^{j(k)}$ can be obtained by the following equation:

$$y^{j(k)} = x^{i(k)} \otimes u_{j(k)}^{i(k)} + b^{j(k)} \qquad (2)$$

where $\otimes$ is the notation of pooling, and $u_{j(k)}^{i(k)}$ partitions the input feature map into a set of non-overlapping rectangles and outputs the maximum or average value for each sub-region. The number of feature maps after pooling remain unchanged.

Additionally, if $L_k$ is the loss layer, then the output value can be obtained on the basis of the following equation:

$$s^{v(k)} = \varphi_k \left( t^{v(k)} z_k + b^{v(k)} \right) \tag{3}$$

where $v(k)$ is the ordinal number of a certain neuron in the loss layer; $s^{v(k)}$ denotes the output value while $b^{v(k)}$ denotes the bias; $t^{v(k)}$ is the weight vector corresponding to the fully connected layer and a neuron in the loss layer; $z_k$ is a vector derived by converting the feature maps of the last pooling layer to a one-dimensional column vector; $\varphi_k(x)$ is the transfer function such as sigmoid and purelin.

Similar to BPNN, the training of parameters of CNN are based on the backward propagation of error and gradient descent algorithm. Furthermore, both of their objectives are minimizing the deviation between the output value and true labels. Typical loss functions (also called performance functions) include Mean Squared Error (MSE) and Cross-entropy Cost.

For simplification, the CNN training pseudocode is shown in Fig. 2. Although CNN and BPNN are both inspired by biological processes, the architecture of CNN is much more complex. Similarities and differences between them are described in Table 1.

```
1   Initialize parameters of the CNN
2   For iteration 1 to N
3       For feature map 1 to M
4           For layer 1 to K
5               If L_k is the input layer
6                   Then input feature map x^{i(k)}
7               Else if L_k is the convolutional layer
8                   Then calculate y^{j(k)} on the basis of Eq. (1)
9               Else if L_k is the pooling layer
10                  Then calculate y^{j(k)} on the basis of Eq. (2)
11              Else if L_k is the fully connected layer
12                  Then convert x^{i(k)} into a one-dimensional column vector z_k
13              Else if L_k is the loss layer
14                  Then calculate the value of s^{v(k)} on the basis of Eq. (3);
15                  Apply a loss function and gradient descent algorithm for backpropagation of error;
16                  Adjust parameters accordingly.
17              End if
18          End for
19      End for
20  End for
21  The training of CNN is completed
```

Fig. 2. The pseudocode of training CNN

Table 1. Comparison between two types of neural networks

|  | CNN | BPNN |
|---|---|---|
| Differences | (i) Reduce the scale of parameters by weight sharing and local connectivity.<br>(ii) Consist of different types of layers, including convolutional layer and pooling layer, etc.<br>(iii) More techniques are required in network training, and parameters are updated slowly due to gradient diffusion. | (i) Neurons are fully connected between adjacent layers.<br>(ii) Comprise a few hidden layers.<br>(iii) Easier training of parameters. |
| Similarities | (i) Feed-forward network.<br>(ii) One-way transmission of input information.<br>(iii) Backpropagation of error.<br>(iv) No connection exists among neurons in the same layer. | |

## 3. Results and Discussion

### 3.1 Analysis of Basic Data

The fleet size of EMU should match the expansion of high-speed railway network. However, it is affected by various factors, such as the development of economy and per capita income. To obtain relatively accurate EMU fleet sizes in following years, we select nine indices as the basis of prediction, including length of high-speed railways in operation, length of railways in operation, passenger traffic of high-speed railways, passenger-kilometers of high-speed railways, passenger traffic of railways, passenger-kilometers of railways, gross domestic product, per capita total income of urban households, number of national railway passenger coaches owned.

According to the Railway Statistic Bulletin (issued by Ministry of Transport of the People's Republic of China) and China Statistical Yearbook (published by National Bureau of Statistics of the People's Republic of China), the fleet size of EMU and real data of nine indices from 2007 to 2015 are shown in Table 2. It should be noted that the real data of nine indices in 2016 have not been published yet, while the real fleet size of EMU in 2016 is 2586 train (one train is equal to eight EMUs). Based on the available data previously mentioned, we apply exponential smoothing method to predict the value of these nine indices from 2016 to 2020. The results are presented in Table 3. On the basis of available documents and reports, we can roughly evaluate the reliability of these predicted values.

(1) According to the Implementation Suggestions for Stimulating the Vitality of Key Social Groups and Increasing the Income of Residents issued by the State Council of China in 2016, the annual income of urban residents in 2020 will be two times as much as in 2010, reaching 42066.84 yuan, which is basically consistent with our predicted value (41222.96 yuan ).

(2) The 13th Five Year Plan points out that GDP will grow at a rate of no less than 6.5% in the planning period and reach over 92.7 trillion yuan in 2020, which is almost the same with the predicted value, namely, 92.9 trillion yuan.

(3) Based on the Medium and Long Term Railway Network Plan issued in 2016, the length of railways in operation will rise to 150 thousand km in 2020, including 30 thousand km high-speed railways. The predicted value of these two indices is 152.4 thousand km and 31.9 thousand km, respectively.

(4) With the building of new railways and capacity expansion of existing lines, railway system will provide better service for passengers. It will not only attract some passenger flows from road and air transportation, but also induce new railway transportation demands. Therefore, it is reasonable that passenger traffic volume and passenger-kilometers of railways keep a steady growth. Expanding railway network definitely requires more transportation equipments to fully utilize the capacity. Thus, the number of national railway passenger coaches owned will keep increasing during the following years.

Table 2. Real data of nine indices and EMU fleet size from 2007 to 2015

| Year | Fleet size of EMU (train) | Length of high-speed railways in operation (thousand km) | Length of railways in operation (thousand km) | Passenger traffic of high-speed railways (thousand persons) | Passenger-kilometers of high-speed railways (billion passenger-km) | Passenger traffic of railways (thousand persons) | Passenger-kilometers of railways (billion passenger-km) | Gross domestic product (billion yuan) | Per capita total income of urban households | Number of national railway passenger coaches owned (coach) |
|---|---|---|---|---|---|---|---|---|---|---|
| 2007 | 105 | 0 | 78 | 0 | 0 | 1356700 | 721.63 | 27023.23 | 14908.61 | 44243 |
| 2008 | 176 | 0.67 | 79.7 | 7340 | 1.56 | 1461930 | 777.86 | 31951.55 | 17067.78 | 45076 |
| 2009 | 285 | 2.70 | 85.5 | 46510 | 16.22 | 1524510 | 787.89 | 34908.14 | 18858.09 | 49354 |
| 2010 | 551 | 5.13 | 91.2 | 133230 | 46.32 | 1676090 | 876.22 | 41303.03 | 21033.42 | 50391 |
| 2011 | 849 | 6.60 | 93.2 | 285520 | 105.84 | 1862260 | 961.23 | 48930.06 | 23979.2 | 54731 |
| 2012 | 1083 | 9.36 | 97.6 | 388150 | 144.61 | 1893370 | 981.23 | 54036.74 | 26958.99 | 57721 |
| 2013 | 1308 | 11.03 | 103.1 | 529620 | 214.11 | 2105970 | 1059.56 | 59524.44 | 26467 | 56841 |
| 2014 | 1712 | 16.46 | 111.8 | 703780 | 282.5 | 2304600 | 1124.19 | 64397.4 | 28843.9 | 60629 |
| 2015 | 2206 | 19.84 | 121 | 961390 | 386.34 | 2534840 | 1196.06 | 68550.58 | 31194.8 | 67706 |

Table 3. Predicted value of nine indices from 2016 to 2020

| Year | Length of high-speed railways in operation (thousand km) | Length of railways in operation (thousand km) | Passenger traffic of high-speed railways (thousand persons) | Passenger-kilometers of high-speed railways (billion passenger-km) | Passenger traffic of railways (thousand persons) | Passenger-kilometers of railways (billion passenger-km) | Gross domestic product (billion yuan) | Per capita total income of urban households | Number of national railway passenger coaches owned (coach) |
|---|---|---|---|---|---|---|---|---|---|
| 2016 | 21.44 | 126.2 | 1208526 | 488.35 | 2735398 | 1252.66 | 73587.14 | 33141.85 | 71979.22 |
| 2017 | 24.06 | 132.7 | 1494775 | 605.75 | 2941246 | 1314.52 | 78428.07 | 35162.13 | 76704.03 |
| 2018 | 26.68 | 139.3 | 1811703 | 735.73 | 3147094 | 1376.38 | 83269 | 37182.4 | 81428.84 |
| 2019 | 29.30 | 145.9 | 2159310 | 878.27 | 3352943 | 1438.23 | 88109.92 | 39202.68 | 86153.65 |
| 2020 | 31.92 | 152.4 | 2537595 | 1033.39 | 3558791 | 1500.09 | 92950.85 | 41222.96 | 90878.46 |

## 3.2 Construction of CNN

Firstly, we define nine indices we selected as virtual pixel points and generate a feature map with a size of $3\times3$. Then, we set three $2\times2$ convolutional kernels and three $2\times2$ pooling filters. In this manner, the structure of a simple CNN can be depicted as Fig. 3.

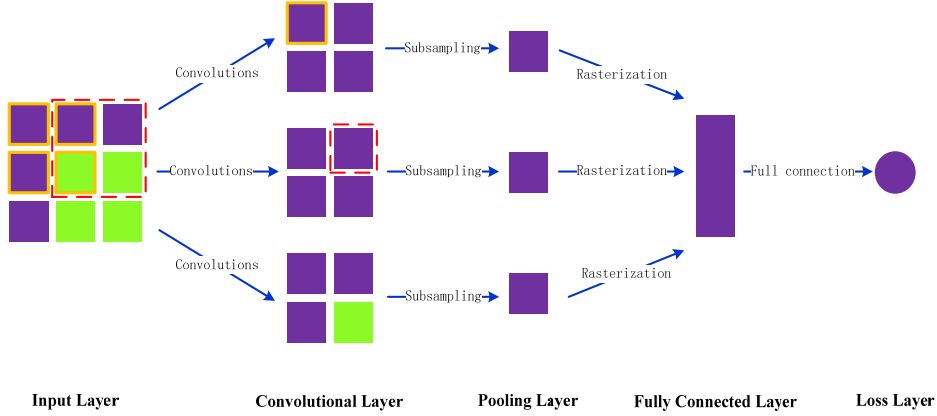

Fig. 3. The architecture of a simple CNN

For this CNN, sigmoid is adopted as the non-linear function for convolutional layer, and average pooling is applied in sub-sampling process without adding bias. For simplicity, the loss function is defined as follows:

$$net.e_m = \frac{1}{2}(net.p_m - net.a_m)^2 \tag{4}$$

where $net.p_m$ represents the predicted value in year $m$, and $net.a_m$ denotes the real fleet size of EMU in the same year. For instance, $net.a_1$ denotes the real fleet size of EMU in 2007, and $net.a_9$ is the size in 2015. We set the learning factor to 0.5, and set the maximum number of iterations to 100. As there are huge differences among the data of different indices, initial data is normalized so as to improve prediction accuracy. The normalization formula is expressed by:

$$net.d'_m = \frac{net.d_m - net.d_{\min}}{net.d_{\max} - net.d_{\min}} \tag{5}$$

where $net.d_m$ denotes the real data of a certain index in year $m$; $net.d_{\max}$ represents the maximum value of this index we collected; $net.d_{\min}$ is the minimum value and $net.d'_m$ is the data after normalization.

## 3.3 Training of CNN

As there are limited available data involved with EMU fleet size, the data from 2007 to 2015 are used for training the network. Computation experiment is carried out on a 2.20 GHz Intel(R) Core (TM) i5-5200U computer with 4.0 GB of RAM, using commercial software MATLAB (R2013b). The computation time of training process is about 3.3 seconds. When the training is completed, parameters of weight vector, convolutional kernels and bias are shown in Table 4.

Table 4. Major parameters of CNN after training

| Kernel Number | Parameters of Convolutional Kernel | Bias of Convolutional Layer |
|---|---|---|
| Kernel 1 | $\begin{bmatrix} -0.5831 & 0.3488 \\ -0.2483 & 0.1905 \end{bmatrix}$ | -0.1579 |
| Kernel 2 | $\begin{bmatrix} 0.8984 & -0.0949 \\ 0.5874 & 0.9605 \end{bmatrix}$ | -1.4013 |
| Kernel 3 | $\begin{bmatrix} 0.1974 & -0.0220 \\ -0.1205 & -0.0415 \end{bmatrix}$ | -0.1631 |
| Weight Vector | | Bias of Loss Layer |
| $\begin{bmatrix} -0.1831 & 2.0049 & -0.2069 \end{bmatrix}$ | | -0.2678 |

As the loss layer of this network contains only one neuron, the MSE of training samples (from 2007 to 2015) can be calculated by the following formula:

$$net.MSE(n) = \frac{1}{9}\sum_{m=1}^{9}(net.p_m - net.a_m)^2 \qquad (6)$$

where $n$ denotes iteration number. The change of MSE in 100 iterations is described in Fig. 4.

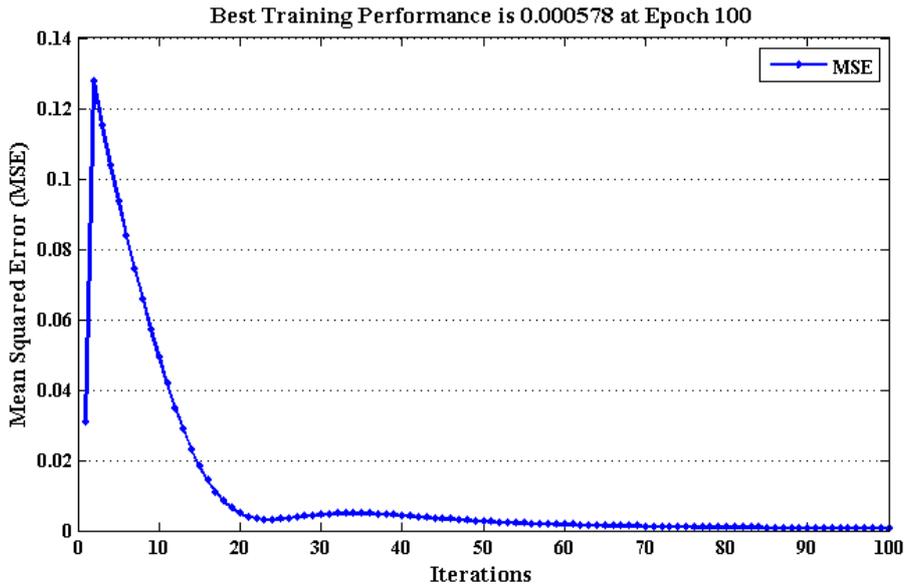

Fig. 4. The behavior of MSE curve in 100 iterations (CNN)

As shown in Fig. 4, the MSE value rapidly increases to 0.1278 at epoch 2 and then drops significantly to 0.004863 at epoch 20. The reason lies in the adoption of stochastic gradient descent algorithm (often shortened in SGD), i.e., adjusting parameters for each training sample. In fact, differences of gradient descent directions truly exist among a number of samples. In other words, the fitting error of a sample may increase when adjusting network parameters based on the error of another. In this case, the MSE of training samples may increase in iteration. Note that, during the next 80 iterations, MSE is in a basically steady status and reaches approximately 0.000578 at the end of iteration, which indicates that the performance of this network can no longer be improved significantly, i.e., no further training is needed.

### 3.4 Analysis of Predicted Results of CNN

Upon the termination of training, we input the data of nine indices from 2016 to 2020 to the trained CNN, and the predicted values from 2016 to 2020 are 2518 trains, 2776 trains, 2973 trains, 3118 trains and 3219 trains respectively.

Note that, the predicted value in 2016 (2518 train) is quite close to the real fleet size (2586 train), which demonstrates that the CNN we trained is of satisfactory generalization ability. To further evaluate the performance of this CNN, we analyze in-depth the fitting error, which is shown in Table 5. In Table 5, error percentage in 2007 and 2008 are -101.9% and -31.25% respectively (in excess of 30%), while the accuracy is greatly improved since 2009 (which is -3.51%), and decreases to 1.04% in 2015. Furthermore, the blue solid line lies close to the red dash line from year 2007 to 2015 (see Fig. 5), which implies that the fitted value basically reflects the changing of EMU fleet size over the past nine years. Thus, the predicted values (from 2016 to 2020) are trustworthy. Under the background of railway network expansion, the EMU fleet size will reach 3219 train in 2020.

Table 5. Fitting error of EMU fleet size (CNN)

| Year | 2007 | 2008 | 2009 | 2010 | 2011 | 2012 | 2013 | 2014 | 2015 |
| --- | --- | --- | --- | --- | --- | --- | --- | --- | --- |
| Actual Value | 105 | 176 | 285 | 551 | 849 | 1083 | 1308 | 1712 | 2206 |
| Fitted value | -2 | 121 | 275 | 522 | 865 | 1106 | 1371 | 1793 | 2229 |
| Error | -107 | -55 | -10 | -29 | 16 | 23 | 63 | 81 | 23 |
| Error Percentage | -101.9% | -31.25% | -3.51% | -5.26% | 1.88% | 2.12% | 4.82% | 4.73% | 1.04% |

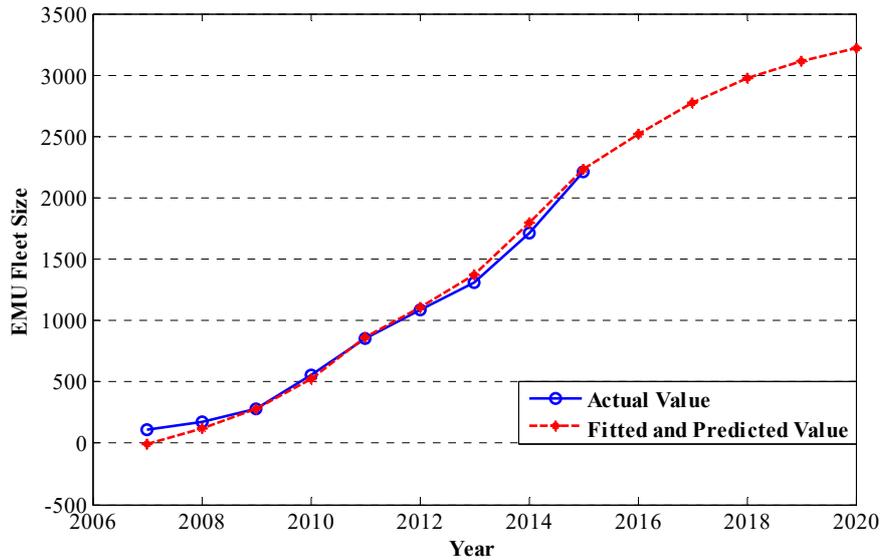

Fig. 5. Curves of EMU fleet size (CNN)

## 3.5 Prediction of EMU Fleet Size Based on BPNN

To compare the performance of CNN and BPNN in prediction, we construct a BPNN using MATLAB (R2013b). This BPNN consists of one input layer, one hidden layer and one output layer. The input layer comprises nine neurons, representing nine indices we selected. The hidden layer is composed of eight neurons, the number of which is determined by trial and error. We adopt sigmoid function as the non-linear function of the hidden layer. The output layer contains only one neuron, and purelin function is chosen as the transfer function. In addition, we implement traingd function to train parameters and adopt learngd as learning function (learning rate is set to 0.5). The computation time is no more than one seconds and the training performance is depicted in Fig. 6.

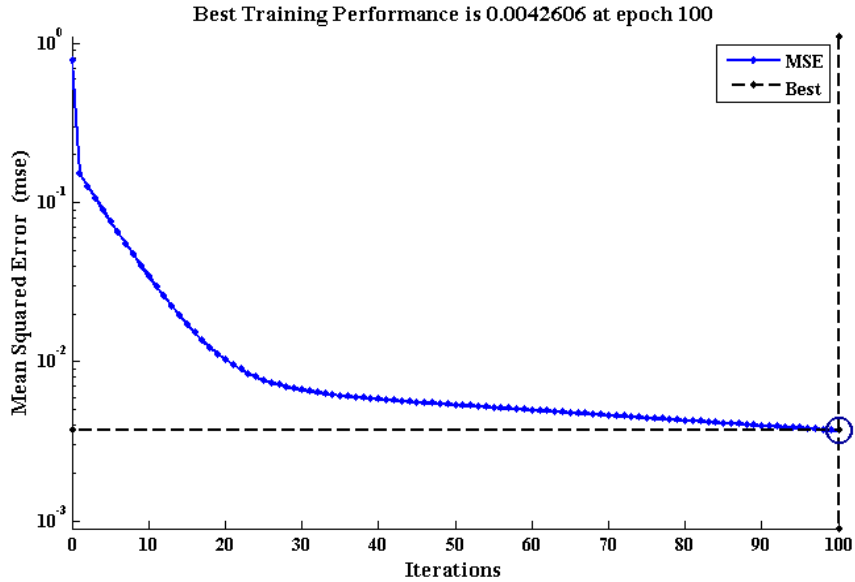

Fig. 6. The behavior of MSE curve (BPNN)

Table 6. Fitted value of EMU fleet size based on BPNN

| Year | 2007 | 2008 | 2009 | 2010 | 2011 | 2012 | 2013 | 2014 | 2015 |
|---|---|---|---|---|---|---|---|---|---|
| Actual Value | 105 | 176 | 285 | 551 | 849 | 1083 | 1308 | 1712 | 2206 |
| Fitted value | -49 | 38 | 378 | 719 | 877 | 1270 | 1318 | 1716 | 1974 |
| Error | -154 | -138 | 93 | 168 | 28 | 187 | 10 | 4 | -232 |
| Error Percentage | -146.67% | -78.41% | 32.63% | 30.49% | 3.30% | 17.27% | 0.76% | 0.23% | -10.52% |

In Fig. 6, the value of MSE falls rapidly then slowly, and finally decreases to 0.0042606 after 100 iterations, which is inferior to that of CNN (0.000578). The predicted value from 2016 to 2020 are 2204 trains, 2543 trains, 2676 trains, 2675 trains and 2650 trains respectively. And the predicted value in 2016 (2204 train) is much smaller than real EMU fleet size (2586 train), which indicates that the CNN after training is of better generalization ability.

From Table 6, the error percentages in 2007 and 2008 are -146.67% and -78.41% respectively, larger than that of CNN. The sum of absolute values of error percentage from 2007 to 2015 reaches 320.28% (see Table 6), while that of CNN is 156.51%. Apparently, the latter is much better. The curves of real fleet size (blue solid line) and predicted fleet size based on BPNN (red dash line) are shown in Fig. 7. To some extent, the fitting performance is not satisfactory. To summarize, CNN can serve as a better aid in EMU fleet size prediction.

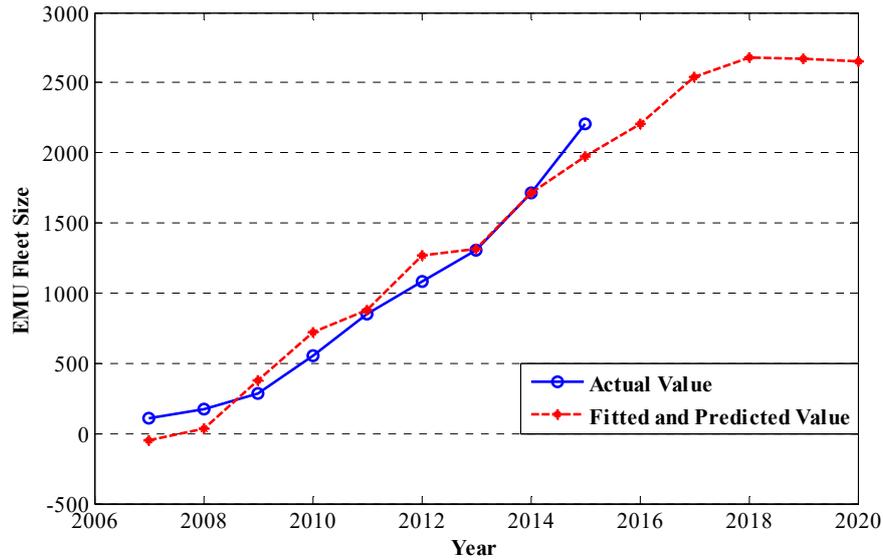

Fig. 7. Curves of EMU fleet size (BPNN)

## 4. Conclusion

In this paper, we not only introduce the basic architecture and theory of CNN, but also make comparisons between CNN and BPNN. It is found that CNN features weight sharing and local connectivity, while more techniques are required in network training. Moreover, nine indices involved with EMU are converted into virtual pixel points so as to generate a feature map with a size of $3\times 3$, and the real data of these indices are analyzed and normalized. In addition, a CNN and a BPNN are constructed and trained to predict the fleet size of EMU from 2016 to 2020 respectively. The results show that the CNN is superior to the BPNN both in generalization ability and fitting accuracy, which illustrates that CNN can be effectively implemented in the prediction of EMU fleet size.

As EMU trains came into service in China since 2007, there is limited available data for our study. Thus, we input nearly all the data we collected to train networks, except for the newly published data of EMU fleet size in 2016, which is used to evaluate the generalization ability of networks. As there are different types of EMU train in operation, researchers can select more indices to predict the fleet size of a certain type of EMU train. We identify this a promising area for future research.